\documentclass{sig-alternate}

\usepackage{algorithm}
\usepackage{algorithmic}
\usepackage{multirow}
\usepackage{subfig}

\begin{document}

\title{Integrate Document Ranking Information into Confidence Measure Calculation for Spoken Term Detection}
\author{
Quan Liu, Wu Guo, Zhen-Hua Ling \\
University of Science and Technology of China, Hefei, China \\
\em quanliu@mail.ustc.edu.cn, guowu@ustc.edu.cn, zhling@ustc.edu.cn\\
}

\maketitle
\begin{abstract}
This paper proposes an algorithm to improve the calculation of confidence measure for spoken term detection (STD).
Given an input query term, the algorithm first calculates a measurement named \textbf{document ranking weight} for each document in the speech database to reflect its relevance with the query term  by summing all the confidence measures of the hypothesized term occurrences in this document.
The confidence measure of each term occurrence is then re-estimated through linear interpolation with the calculated document ranking weight to improve its reliability by integrating document-level information.
Experiments are conducted on three standard STD tasks for Tamil, Vietnamese and English respectively.
The experimental results all demonstrate that the proposed algorithm achieves consistent improvements over the state-of-the-art method for confidence measure calculation.
Furthermore, this algorithm is still effective even if a high accuracy speech recognizer is not available, which makes it applicable for the languages with limited speech resources.
\end{abstract}




\section{Introduction}
Spoken term detection (STD) is a task designed for efficient keyword search (given text query) in a speech databases, and plays a central role in information management and speech retrieval \cite{mamou2007vocabulary,fiscus2007results,kohler2008spoken,mangu2014efficient}.
State-of-the-art STD approaches include two subsystems.
The first one is an automatic speech recognizer (ASR), which is used to transcribe the spoken utterances into text.
The text transcriptions contain all the possibly recognized words with corresponding posterior probabilities \cite{jiang2005confidence,mamou2007vocabulary,mangu2014efficient}. The posterior probability as been one typical \textbf{confidence measure} plays a central role in keyword searching.
The second subsystem is a keyword searcher which returns the results of term detection for each query term according to the decoded transcriptions.
Formally, in STD applications, a confidence measure (CM) is defined to represent the reliability of each detected term occurrence, which is usually estimated by the recognizer \cite{jiang2005confidence,mamou2007vocabulary}.
Relying on the confidence measure, the final term detection results could be obtained by threshold-based recall.
However, when only limited training resources are available for building the ASR system,
the accuracy of the recognizer and the reliability of the confidence measure are relatively low, which makes it difficult to find correct query results in the speech database.

This paper focuses on the calculation of confidence measure for STD when the speech recognizer has been built.
In this situation, a one-pass retrieval candidate set can be obtained for each query.
Each candidate contains the term occurrence location information and the corresponding confidence measure.
The baseline system of this paper could then be evaluated on it directly by conducting standard score normalization and final decision \cite{mamou2013system,mangu2014efficient}.
\textbf{To improve the reliability of term occurrences}, some recent efforts have attempted to do this work and have achieved some improvements on STD task.
In \cite{li2012novel,lee2011improved}, the confidence measure of query occurrence is re-estimated based on the context consistency information.
\cite{soto2014comparison} proposed a two-stage cascaded machine learning approach for rescoring keyword search outputs for low resource languages.
\cite{hout2014comb} proposed a modified logistic regression strategy for term detection optimization.
Discriminative score normalization method was introduced to normalize confidence measures through discriminative modeling \cite{pham2014discriminative}.
Moreover, another method was proposed in \cite{lee2014improved} to employ extra acoustic features for getting a better confidence measure.

However, all these methods fail to utilize long-term contexts at document or topic level, which has been proved to be useful for some other information retrieval (IR) tasks \cite{ponte1998language,zhai2004study}.
Clustering and latent topic models have also gained improvements over traditional vector space models for IR \cite{wei2006lda,chen2009latent}.
Besides, the well known PageRank algorithm considers the hyperlink between every two pages and computes a converged importance score for each page \cite{brin1998anatomy}.
Inspired by these work, this paper proposes to integrate document ranking information into the calculation of confidence measures of term occurrences for spoken term detection.
The document ranking information is defined to be the \textbf{topic relevance} between the document and query term. For each query term, there are some documents tend to be more related to it because they are of a similar topic. When examining the accuracy of STD results, those topic-related documents tend to contain more correct hits.
In detail, this information is quantized as a ranking weight for each document in this paper.
Based on the one-pass retrieval candidates for a specific query term, we first sum up the confidence measures of all term occurrences in each document.
The document ranking weights are then estimated by normalizing these sums and are further integrated into the original confidence measures through linear interpolation.
Experiments on three standard STD tasks demonstrate the effectiveness of our proposed method.

For the rest of this paper, we will describe the related works of this paper in Section \ref{sec:related}. The proposed algorithm for confidence measure calculation will be presented in Section \ref{sec:main}. Section \ref{sec:experm-set} and \ref{sec:experm-res} are the experimental setup and results on three standard  STD tasks. Finally, we will conclude our work in Section \ref{sec:conclusion}.

\section{Related Work}
\label{sec:related}
There are some other work attempted to utilize long-term contexts for STD.
In \cite{chiu2013using}, they improved term detection performance based on the word burstiness in spoken conversational corpora.
More recently, \cite{wintrode2014can,richards2014using} took advantage of word repetition to improve spoken term detection, having observed the phenomenon of word repetition within single documents. They leveraged the burstiness of keywords by taking the most confident keyword hypothesis in each document and interpolating with lower scoring hits.
Although they had designed an effective method to determine the inter coefficients in their experiments, they focussed on intra-document term repetition, without paying attention to the inter-document contexts, e.g. the document ranking information used in this paper.
The work in \cite{konno2013high} is very similar to us since they also gave a high priority to the candidate segments that are included in highly ranked documents. However, they proposed to calculate the position dependent document weights recursively.
This paper calculates document ranking weights in a more easier way and considers the inter document ranking information.
In this paper, we will rank all documents in the speech database according to their relevance with a specific query term and incorporate such document ranking information into the calculation of  confidence measures.

\section{Proposed Method}
\label{sec:main}
\begin{algorithm}[t]
\footnotesize
\caption{\small Calculate Document Ranking Weights Given a Query Term}
\textbf{Input:} The set of one-pass retrieval candidates given query term $t$. \\
\textbf{Output:} The document ranking weights for all documents in the database.\\
\textbf{Main procedure:}
\begin{enumerate}
   \item \textbf{Document Clustering}\\
   Cluster the documents in all the hypothesized occurrences of term $t$ by summing all the confidence measures in each document $d$:
   \\
    \begin{equation}\label{rank-1}
        S_{d}(t) = \sum_{O_{i} \in d} \mathrm{CM}_{\mathrm{base}}(t|O_{i},d),
    \end{equation}
   where $S_{d}(t)$ can be viewed as the occurrence possibility of term $t$ in document $d$.
   The maximum score $S_{\textrm{max}}(t)$ for term $t$ can also be obtained if we traverse all the documents.
   \begin{equation}\label{rank-2}
        S_{\textrm{max}}(t) = \mathop{\max}_{d \in \textrm{all documents}}{S_{d}(t)}.
    \end{equation}

   \item \textbf{Document Ranking} \\
   The ranking weight $W_{d}(t)$ for each document is calculated using the ``relative-to-max'' method, which is obtained by dividing $S_{d}(t)$ by $S_{\textrm{max}}(t)$:
   \\
   \begin{equation}\label{rank-3}
        W_{d}(t) = S_{d}(t) / S_{\textrm{max}}(t).
    \end{equation}
\end{enumerate}
\textbf{End} \\
\label{algorithm-core}
\end{algorithm}

For an input query term, a set of one-pass retrieval candidates in the speech database is firstly generated following the conventional STD approach.
Each term detection occurrence commonly contains location information and a confidence measure, while the location information usually includes the located document name (or ID), start time and duration time.
For example, for term $t$, we use $O_{i}$ to represent the location information of the $i$-th detection occurrence of term $t$.
If the location information indicates that this occurrence candidate belong to document $d$, then the confidence measure of the $i$-th term detection occurrence confidence measure can be denoted as $\mathrm{CM}_{\mathrm{base}}(t|O_{i}, d)$.
We use subscript ``\textrm{base}'' to emphasis that this measure is obtained from the one-pass retrieval candidate set.
The confidence measure is designed to describe the reliability of a detected term occurrence, i.e., a correct query hit is expected to have a high confidence measure.
However, when the ASR subsystem performs poorly, there may be many false alarms with high confidence measure as well as correct candidates with low confidence measure.

Based on the idea we have described in the introduction section, we propose to use document ranking information to improve the calculation of confidence measures.
The algorithm to estimate the document ranking weight $W_{d}(t)$ for a input term $t$ is described in Algorithm \ref{algorithm-core}.
After the calculation of document ranking weights, we re-estimate the confidence measure of each occurrence 
by combining the original one with the ranking weight of the document it belongs to. In this work, a linear interpolation is adopted as
\begin{equation}\label{rank-4}
    \mathrm{CM}_{\mathrm{new}}(t|O_{i},d) = \alpha W_{d}(t) + (1-\alpha) \mathrm{CM}_{\mathrm{base}}(t|O_{i},d),
\end{equation}
where the interpolation coefficient $\alpha$ for interpolation is consistent for all query terms, and it can be tuned using a development set. In short, the algorithm of confidence re-estimation can be divided into three steps, i.e., document clustering, document ranking and confidence re-estimation.

%

\section{EXPERIMENTAL SETUP}
\label{sec:experm-set}
\subsection{Data Set and Evaluation Condition}
The experiments were conducted using three standard spoken term detection tasks, the STD 2006 English conversational telephone speech (CTS) evaluation set, the OpenKWS 2013 Vietnamese and the OpenKWS 2014 Tamil development sets\footnote{http://www.nist.gov/itl/iad/mig/openkws.cfm}.
The English CTS evaluation set included about 3 hours of speech, and the keyword set consisted of 411 keywords. The development sets of Vietnamese and Tamil included about 10 hours of speech respectively. The evaluation keyword set for Vietnamese consisted of 4065 keywords, with 901 of those keywords appearing in the development set and being used in our experiments.
For the Tamil task, we used the kwlist3 keyword set supplied by IBM, which consisted of 2375 keywords. The intention of using three tasks was to evaluate the proposed algorithm using three very different languages, with different ASR accuracy, different amounts of training data and with variations in the sizes of keyword sets.
The evaluation criterion used in the experiments was the Actual Term Weighted Value (ATWV) defined by NIST, using a cost function of the false alarm probability P(FA) and P(Miss), averaged over a set of queries\footnote{http://www.itl.nist.gov/iad/mig/tests/std/2006/docs/std06-evalplan-v10.pdf}.

\subsection{Automatic Speech Recognizer}
Our ASR engines were built using the DNN-HMM based acoustic modeling, which is the state-of-the-art approach for speech recognition \cite{seide2011conversational}.

For the English task, 309 hours of Switchboard speech were used to train the acoustic model, and the transcriptions of these speech files were used to train a 3-gram language model. The cross entropy criterion was used to train the DNN models. The word accuracy (ACC) of the ASR system on the evaluation set was \textbf{77.67\%}.

For the Vietnamese recognizer, two approaches were adopted to prevent the over-fitting problem in DNN training since the training corpus contains only about 70 hours of speech.
The first approach was cross-lingual training, where we used a DNN model acquired from 1000 hours of Chinese CTS data to initialize the Vietnamese DNN parameters.
Furthermore, the rectified linear unit (ReLU) activation function was used to replace the sigmoid function in the DNN model.
The transcripts of the Vietnamese training files were then used to train a 2-gram language model. A word ACC of \textbf{45.76\%} was achieved on the development set.
The strategy employed for the Tamil ASR engine was similar to that used for Vietnamese.
The only difference was that the sequence training algorithm was applied in the DNN training for Tamil. A word ACC of \textbf{31.03\%} was achieved on the development set.

\subsection{STD Indexer and Keyword Searcher}
We designed a toolkit named iSTD to build our keyword search subsystem for STD. We followed the work in \cite{mamou2007vocabulary,mangu2014efficient} to construct the inverted index based on confusion networks.
The term occurrence candidates were then found by keyword searching on the inverted index. The confidence re-estimation algorithm proposed in this paper was also integrated into this toolkit.

\section{Experimental Results}
\label{sec:experm-res}
\subsection{Effectiveness of Document Ranking}
\begin{figure}[tb]
\begin{minipage}[b]{1.0\linewidth}
  \centering
  \vspace{-2mm}
  \includegraphics[width=4.1cm]{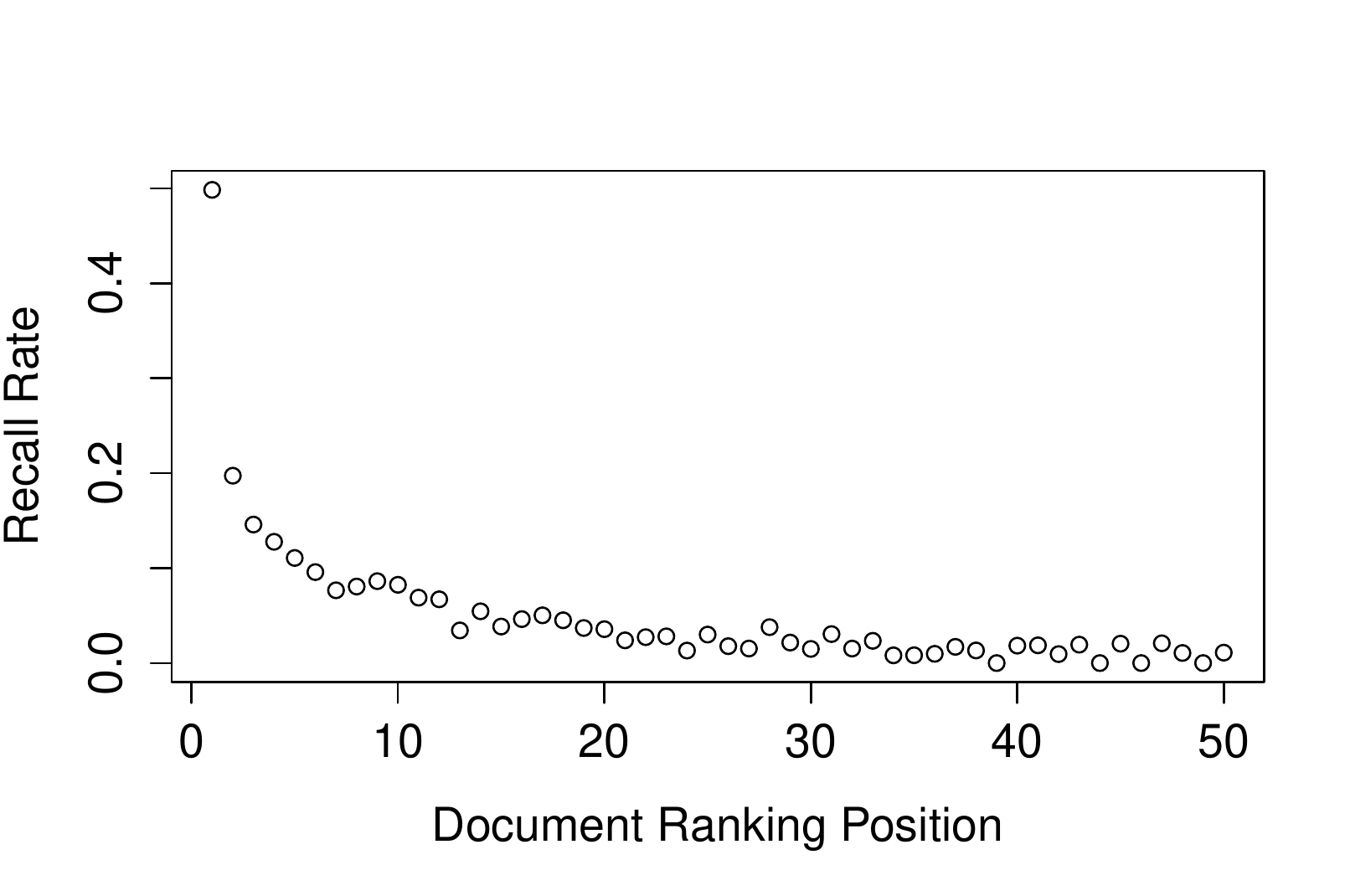}
  \includegraphics[width=4.1cm]{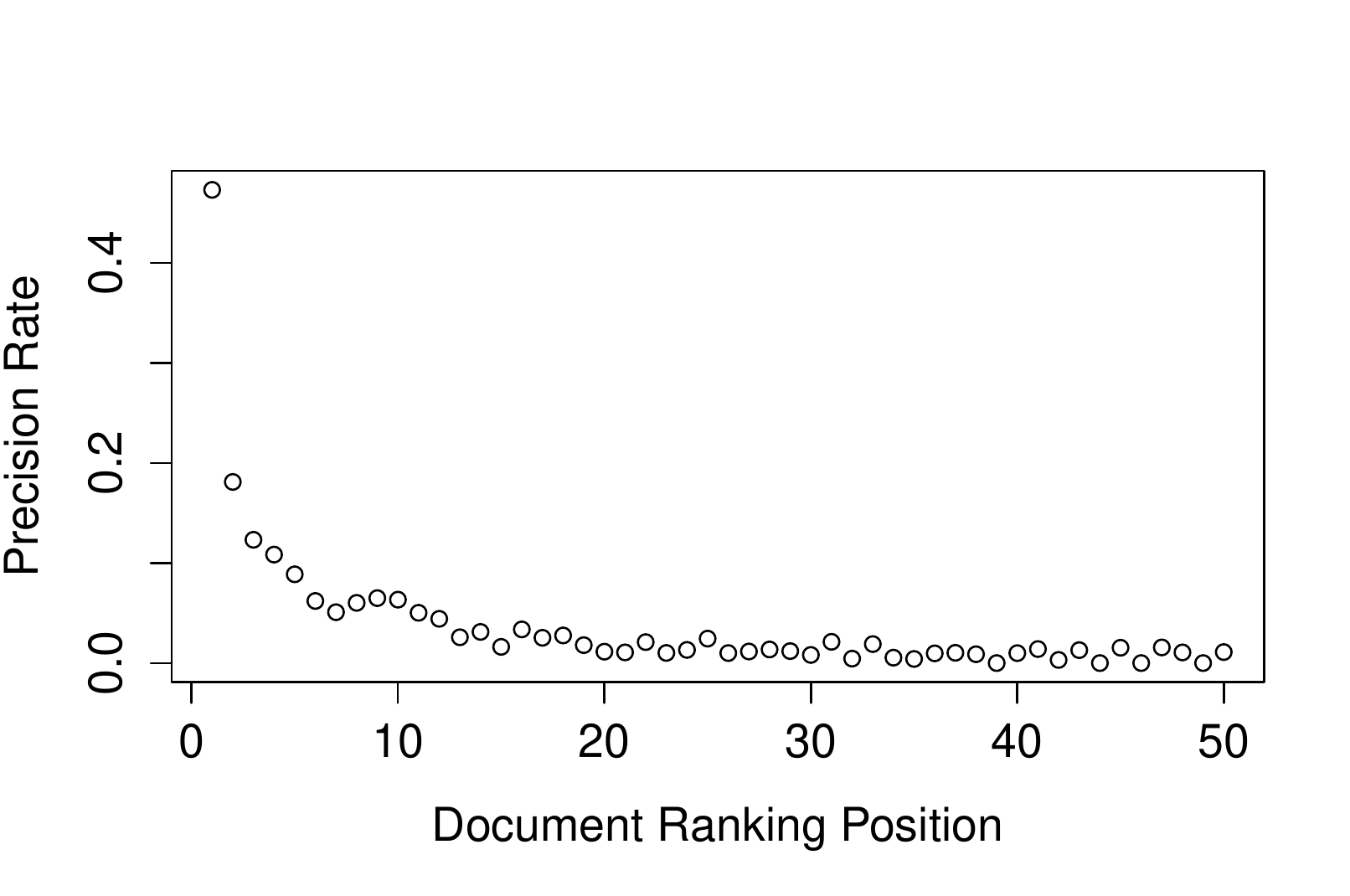}
  \vspace{-3mm}
\end{minipage}
\caption{Correlation Curve based on Document Ranking.}
\label{fig:correlation}
\end{figure}

In order to validate the rationality of applying the document ranking information to STD tasks, we examined the relationship between the performance of term detection and the document ranking positions.
Here, the document ranking positions were derived by sorting all documents in descending order of the weights calculated following Algorithm \ref{algorithm-core}.
Figure \ref{fig:correlation} shows the correlation curve for the aforementioned Vietnamese STD task.
The results were obtained by averaging over 901 query keywords.
The correlation curves reveal that the documents with high document ranking weights usually have high  precision and recall of term detection.

In addition, we calculated the Spearman rank correlation coefficient between the two performance measurement of term detection and the document ranking weights on the three STD tasks. The results are given in Table \ref{tab:spearman} and shows the existence of high correlations.
All these results indicate that the document ranking information is strongly correlated with the STD performance and it is reasonable to integrate it into the calculation of confidence measures for the term detection.

\begin{table}[bth]
\centering
\caption{Spearman correlation for three STD Tasks.}
\begin{tabular}[htb]{|c|c|c|c|}
  \hline
  \multicolumn{2}{|c|}{STD Task} & \multicolumn{2}{c|}{Spearman Correlation} \\\cline{1-4}
  Language & ACC & Precision-Rank & Recall-Rank  \\\hline\hline
  English & 78\% & 0.93 & 0.74 \\\hline
  Vietnamese & 46\% & 0.74 & 0.73 \\\hline
  Tamil & 31\% & 0.70 & 0.68 \\\hline
\end{tabular}
\label{tab:spearman}
\end{table}

\subsection{Results of Tuning Interpolation Coefficients}
\begin{figure}[htb]
\begin{minipage}[b]{1.0\linewidth}
  \centering
  \centerline{\includegraphics[width=7.5cm]{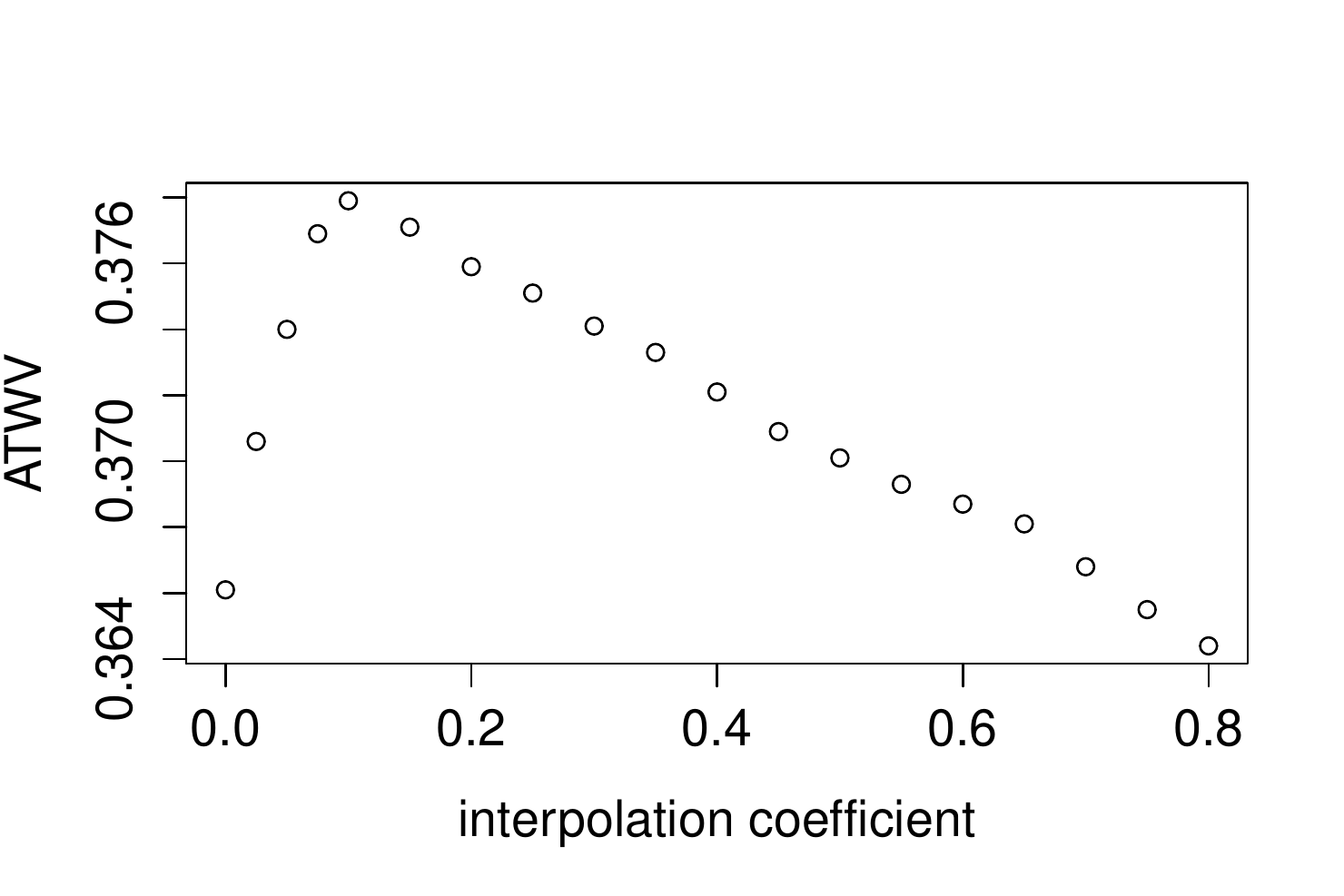}}
\end{minipage}
\caption{Effect of different interpolation coefficient.}
\label{fig:interpolation}
\end{figure}

The interpolation coefficient $\alpha$ in (4) controls the balance between the document weights and the baseline confidence measures for a specific query term. To explore its practical effcets, the ATWVs on the development set of the Vietnamese STD task versus different interpolation coefficients were depicted in Fig. \ref{fig:interpolation}.
We can see that a reasonable choice for $\alpha$ is within the range 0.05 to 0.4.
In the next section, experimental results will be presented for different tasks,
where $\alpha$  was tuned on the development and set to be 0.05, 0.1 and 0.15 for Tamil, Vietnamese and English respectively.

\subsection{Results of STD Tasks}
We compared the proposed confidence measure re-estimation algorithm with the baseline system for the three STD tasks. The baseline system directly adopted the ASR posterior score as the confidence measure for each query term. Keyword-specific threshold was applied for all systems as the final decision recall method \cite{wang2014depth}. Experimental results are listed in Table \ref{figres:cores}.
We can see that the proposed confidence re-estimation approach achieves consistent improvements for all the three typical speech retrieval tasks. Considering the amount of training data available in these three tasks, the results in Table \ref{figres:cores} also indicate that the proposed confidence re-estimation method is neither language-dependent, nor sensitive to the amounts of training resources.

\begin{table}
\centering
\caption{Term Detection Results for Three Tasks (ASR recognition accuracy: English=78\%, Vietnamese=46\%, Tamil=31\%).}
\begin{tabular}[htb]{|c|c|c|c|}
  \hline
  Language & Confidence & ATWV & P(Miss)
  \\\hline\hline
  \multirow{2}{*}{English} & Baseline & 0.8064 & 0.142\\\cline{2-4}
  & Proposed & \textbf{0.8182 (+1.5\%)} & \textbf{0.119}\\\hline
  \multirow{2}{*}{Vietnamese} & Baseline & 0.3661 & 0.583\\\cline{2-4}
  & Proposed & \textbf{0.3779 (+3.2\%)} & \textbf{0.565} \\\hline
  \multirow{2}{*}{Tamil} & Baseline & 0.2785 & 0.661\\\cline{2-4}
   & Proposed & \textbf{0.2934 (+5.4\%)} & \textbf{0.626}\\\hline
\end{tabular}
\label{figres:cores}
\end{table}


\section{CONCLUSIONS}
\label{sec:conclusion}
This paper has presented an algorithm to improve the calculation of confidence measures for spoken term detection.
Inspired by the PageRank algorithm and the application of language models in the text information retrieval area, we propose to integrate the document ranking information into the calculation of confidence measures for term occurrences.
The document ranking information indicates the topic relevance between each document and the query term, while topic-related documents are expected to contain more correct hits.
Experiments on three standard STD tasks demonstrate the effectiveness of this algorithm by introducing document ranking information.

%
%
\bibliographystyle{abbrv}
\bibliography{sigproc}  

\begin{thebibliography}{10}

\bibitem{brin1998anatomy}
S.~Brin and L.~Page.
\newblock The anatomy of a large-scale hypertextual web search engine.
\newblock {\em Computer networks and ISDN systems}, 30(1):107--117, 1998.

\bibitem{chen2009latent}
B.~Chen.
\newblock Latent topic modelling of word co-occurence information for spoken
  document retrieval.
\newblock In {\em Proc. ICASSP}, pages 3961--3964. IEEE, 2009.

\bibitem{chiu2013using}
J.~Chiu and A.~I. Rudnicky.
\newblock Using conversational word bursts in spoken term detection.
\newblock In {\em Proc. INTERSPEECH}, pages 2247--2251, 2013.

\bibitem{fiscus2007results}
J.~G. Fiscus, J.~Ajot, J.~S. Garofolo, and G.~Doddingtion.
\newblock Results of the 2006 spoken term detection evaluation.
\newblock In {\em Proc. SIGIR}, volume~7, pages 51--57, 2007.

\bibitem{jiang2005confidence}
H.~Jiang.
\newblock Confidence measures for speech recognition: A survey.
\newblock {\em Speech communication}, 45(4):455--470, 2005.

\bibitem{kohler2008spoken}
J.~Kohler, M.~Larson, F.~de~Jong, W.~Kraaij, and R.~Ordelman.
\newblock Spoken content retrieval: Searching spontaneous conversational
  speech.
\newblock In {\em ACM SIGIR Forum}, volume~42, pages 66--75. ACM, 2008.

\bibitem{konno2013high}
K.~Konno, Y.~Itoh, K.~Kojima, M.~Ishigame, K.~Tanaka, and S.-w. Lee.
\newblock High priority in highly ranked documents in spoken term detection.
\newblock In {\em Signal and Information Processing Association Annual Summit
  and Conference (APSIPA), 2013 Asia-Pacific}, pages 1--4. IEEE, 2013.

\bibitem{lee2014improved}
H.-y. Lee, P.-w. Chou, and L.-s. Lee.
\newblock Improved open-vocabulary spoken content retrieval with word and
  subword lattices using acoustic feature similarity.
\newblock {\em Computer Speech $\&$ Language}, 2014.

\bibitem{lee2011improved}
H.-y. Lee, T.-w. Tu, C.-P. Chen, C.-y. Huang, and L.-s. Lee.
\newblock Improved spoken term detection using support vector machines based on
  lattice context consistency.
\newblock In {\em Proc. ICASSP}, pages 5648--5651, 2011.

\bibitem{li2012novel}
H.~Li, J.~Han, T.~Zheng, and G.~Zheng.
\newblock A novel confidence measure based on context consistency for spoken
  term detection.
\newblock In {\em Proc. INTERSPEECH}, 2012.

\bibitem{mamou2013system}
J.~Mamou, J.~Cui, X.~Cui, M.~J.~F. Gales, B.~Kingsbury, K.~Knill, L.~Mangu,
  D.~Nolden, M.~Picheny, B.~Ramabhadran, R.~Schl\"uter, A.~Sethy, and P.~C.
  Woodl.
\newblock System combination and score normalization for spoken term detection.
\newblock In {\em Proc. ICASSP}, pages 8272--8276, 2013.

\bibitem{mamou2007vocabulary}
J.~Mamou, B.~Ramabhadran, and O.~Siohan.
\newblock Vocabulary independent spoken term detection.
\newblock In {\em Proc. SIGIR}, pages 615--622. ACM, 2007.

\bibitem{mangu2014efficient}
L.~Mangu, B.~Kingsbury, H.~Soltau, H.-K. Kuo, and M.~Picheny.
\newblock Efficient spoken term detection using confusion networks.
\newblock In {\em Proc. ICASSP}, pages 7844--7848, 2014.

\bibitem{pham2014discriminative}
V.~T. Pham, H.~Xu, N.~F. Chen, S.~Sivadas, B.~P. Lim, E.~S. Chng, and H.~Li.
\newblock Discriminative score normalization for keyword search decision.
\newblock In {\em Proc. ICASSP}, pages 7078--7082, 2014.

\bibitem{ponte1998language}
J.~M. Ponte and W.~B. Croft.
\newblock A language modeling approach to information retrieval.
\newblock In {\em Proc. SIGIR}, pages 275--281. ACM, 1998.

\bibitem{wang2014depth}
Y.~Proc.~Wang and F.~Metze.
\newblock An in-depth comparison of keyword specific thresholding and
  sum-to-one score normalization.
\newblock In {\em INTERSPEECH}, 2014.

\bibitem{richards2014using}
J.~Richards, M.~Ma, and A.~Rosenberg.
\newblock Using word burst analysis to rescore keyword search candidates on
  low-resource languages.
\newblock In {\em Proc. ICASSP}, pages 7824--7828, 2014.

\bibitem{seide2011conversational}
F.~Seide, G.~Li, and D.~Yu.
\newblock Conversational speech transcription using context-dependent deep
  neural networks.
\newblock In {\em INTERSPEECH}, pages 437--440, 2011.

\bibitem{soto2014comparison}
V.~Soto, L.~Mangu, A.~Rosenberg, and J.~Hirschberg.
\newblock A comparison of multiple methods for rescoring keyword search lists
  for low resource languages.
\newblock In {\em Proc. INTERSPEECH}, 2014.

\bibitem{hout2014comb}
J.~van Hout, L.~Ferrer, D.~Vergyri, N.~Scheffer, Y.~Lei, V.~Mitra, and
  S.~Wegmann.
\newblock Calibration and multiple system fusion for spoken term detection
  using linear logistic regression.
\newblock In {\em Proc. ICASSP}, pages 7188--7192, 2014.

\bibitem{wei2006lda}
X.~Wei and W.~B. Croft.
\newblock Lda-based document models for ad-hoc retrieval.
\newblock In {\em Proc. SIGIR}, pages 178--185. ACM, 2006.

\bibitem{wintrode2014can}
J.~Wintrode and S.~Khudanpur.
\newblock Can you repeat that? using word repetition to improve spoken term
  detection.
\newblock In {\em Proc. ACL}, pages 1316--1325. Association for Computational
  Linguistics, 2014.

\bibitem{zhai2004study}
C.~Zhai and J.~Lafferty.
\newblock A study of smoothing methods for language models applied to
  information retrieval.
\newblock {\em ACM Transactions on Information Systems (TOIS)}, 22(2):179--214,
  2004.

\end{thebibliography}
%
%

\end{document}